\pdfoutput=1

\documentclass[11pt]{article}

\usepackage{EMNLP2022}

\usepackage{times}
\usepackage{latexsym}

\usepackage[T1]{fontenc}

\usepackage[utf8]{inputenc}

\usepackage{microtype}

\usepackage{inconsolata}
\usepackage{mathrsfs}
\usepackage{algorithm}
\usepackage{algpseudocode}
\usepackage{amsmath}
\usepackage{booktabs}
\usepackage{subfigure}
\usepackage{graphicx}
\usepackage{multirow}
\usepackage{arydshln}
\usepackage{float}

\title{Forging Multiple Training Objectives for Pre-trained Language Models via Meta-Learning}

\author{Hongqiu Wu\textsuperscript{\rm 1,\rm 2,$\spadesuit$}, Ruixue Ding\textsuperscript{\rm $\spadesuit$}, Hai Zhao\textsuperscript{\rm 1,\rm 2,\thanks{\; Corresponding author.}}, \\
        {\bf Boli Chen}\textsuperscript{\rm $\spadesuit$}, {\bf Pengjun Xie}\textsuperscript{\rm $\spadesuit$}, {\bf Fei Huang}\textsuperscript{\rm $\spadesuit$}, {\bf Min Zhang}\textsuperscript{\rm 3} \\
        \textsuperscript{\rm 1}Department of Computer Science and Engineering, Shanghai Jiao Tong University \\
        \textsuperscript{\rm 2}Key Laboratory of Shanghai Education Commission for Intelligent Interaction \\
        and Cognitive Engineering, Shanghai Jiao Tong University \\
        \textsuperscript{\rm 3}School of Computer Science and Technology, Soochow University \\
        \textsuperscript{\rm $\spadesuit$}Damo Academy, Alibaba Group \\
        \texttt{wuhongqiu@sjtu.edu.cn}, \texttt{zhaohai@cs.sjtu.edu.cn}, \\
        \texttt{\{ada.drx,boli.cbl,chengchen.xpj,f.huang\}@alibaba-inc.com}, \\
        \texttt{minzhang@suda.edu.cn}}

\begin{document}
\maketitle
\begin{abstract}
Multiple pre-training objectives fill the vacancy of the understanding capability of single-objective language modeling, which serves the ultimate purpose of pre-trained language models (PrLMs), generalizing well on a mass of scenarios. However, learning multiple training objectives in a single model is challenging due to the unknown relative significance as well as the potential contrariety between them. Empirical studies have shown that the current objective sampling in an ad-hoc manual setting makes the learned language representation barely converge to the desired optimum. Thus, we propose \textit{MOMETAS}, a novel adaptive sampler based on meta-learning, which learns the latent sampling pattern on arbitrary pre-training objectives. Such a design is lightweight with negligible additional training overhead. To validate our approach, we adopt five objectives and conduct continual pre-training with BERT-base and BERT-large models, where MOMETAS demonstrates universal performance gain over other rule-based sampling strategies on 14 natural language processing tasks\footnote{\url{https://github.com/gingasan/mometas}}.
\end{abstract}

\section{Introduction}

It is appealing for deep neural language models to generalize well on multiple downstream tasks through large-scale language pre-training, e.g. BERT \citep{DBLP:conf/naacl/DevlinCLT19}, ELECTRA \citep{DBLP:conf/iclr/ClarkLLM20}, DeBERTa \citep{he2021deberta} and GPT \citep{DBLP:conf/nips/BrownMRSKDNSSAA20}. Most pre-trained language models (PrLMs) rely on only one or two pre-training objectives, from Masked Language Modeling (MLM), Next Sentence Prediction (NSP) \citep{DBLP:conf/naacl/DevlinCLT19}, Sentence Order Prediction (SOP) \citep{DBLP:conf/iclr/LanCGGSS20} and Permutation Language Modeling (PLM) \citep{DBLP:conf/nips/YangDYCSL19}. Even though PrLMs are intended for high generalization, studies show that they are not always all-rounded and tend to be particularly weak in some aspects \citep{DBLP:conf/acl/LiZ20, DBLP:conf/emnlp/LiZHWYL20, DBLP:conf/nips/YangDYCSL19}, while an ultimate purpose of a language understanding system is to stand for the nice initialization on a mass of scenarios simultaneously and effectively.

With the birth of more and more pre-training objectives, a number of specific ones beyond are found of great benefit to enhance task-level understanding capability, e.g. contrastive learning \citep{DBLP:conf/emnlp/GaoYC21}, adversarial training \citep{DBLP:journals/corr/abs-2206-12608}, knowledge injection \citep{DBLP:conf/iclr/XiongDWS20}. To enjoy the merits of all worlds and let the model generalize better on more seen or perhaps unseen tasks, there naturally comes a need to combine all these objectives in an organic manner.

However, learning multiple pre-training objectives simultaneously in a single model is challenging \citep{DBLP:conf/icml/ChenBLR18,DBLP:conf/nips/YuK0LHF20}. A well-known issue is negative transfer \citep{DBLP:conf/cvpr/WangDPC19} in which learning well on one objective impairs another. More importantly, the relative significance between all objectives is supposed to be scheduled. For instance, NSP can take little effect on the model due to its simpleness in the mature stage of training. However, it is of great difficulty to heuristically tune such a ratio considering the large amounts of compute to pre-train once. In most cases we tentatively treat all of them equally \citep{DBLP:conf/acl/LiuHCG19,DBLP:conf/acl/LewisLGGMLSZ20}, which makes the learned language representation barely converge to the optimal point and limits the model performance \citep{DBLP:conf/icml/ChenBLR18,DBLP:conf/cvpr/WangDPC19}.

To forge multiple training objectives for PrLMs, this paper presents to learn an optimal sampling strategy so that the more informative objective is more likely to be chosen. The backbone is meta-learning \citep{DBLP:books/sp/98/ThrunP98} and thus we call it \textit{\textbf{M}ulti-\textbf{O}bjective \textbf{META}-\textbf{S}ampler} (\textit{MOMETAS}). In the proposed framework, we redesign the pre-training process into two phases, meta-train and meta-test. The model is trained alternately on one sampled objective at each step during meta-train, while the sampling distribution is then updated during meta-test by measuring the relative contribution of each objective. The training design is lightweight with little additional overhead to guarantee the pre-training efficiency. To validate our approach, we consider five pre-training objectives (e.g. for sentence embedding, knowledge caption, syntactic understanding) and continue to pre-train with BERT-base and BERT-large, where MOMETAS demonstrates universal performance gain over other rule-based sampling strategies on 14 natural language processing tasks.

\section{Related Work}

\subsection{Multiple Pre-training Objectives}

Our work is dedicated to improvement of learning multiple pre-training objectives on a single language model \citep{DBLP:conf/acl/LiuHCG19,DBLP:conf/acl/LewisLGGMLSZ20}. Language pre-training is well-studied in recent years and there are various potential objectives proposed, e.g. to enhance general language representation \citep{DBLP:conf/acl/LewisLGGMLSZ20}, text generation \citep{DBLP:conf/nips/YangDYCSL19,DBLP:conf/nips/00040WWLWGZH19}, sentence embedding \citep{DBLP:conf/emnlp/GaoYC21,DBLP:conf/acl/LiZ20}, dialogue understanding \citep{DBLP:conf/acl/XuZ21,DBLP:conf/emnlp/0002Z21}. MOMETAS is designed to bring them together organically.

Our work is related to balancing training in multi-task networks, e.g. gradient normalization \citep{DBLP:conf/icml/ChenBLR18}, projecting conflicting gradients \citep{DBLP:conf/nips/YuK0LHF20}, weighting training loss based on uncertainty \citep{DBLP:conf/cvpr/KendallGC18}. For PrLMs, it is explored more on fine-tuning \citep{DBLP:conf/icml/Stickland019,DBLP:journals/jmlr/RaffelSRLNMZLL20,DBLP:conf/emnlp/PothPRG21}. In practice, BERT-style pre-training like MLM \citep{DBLP:conf/naacl/DevlinCLT19} establishes self-supervised objectives through certain transformations on text data. From this point of view, our work is similar to reweighting training samples \citep{DBLP:journals/corr/AlainLSCB15,DBLP:conf/icml/RenZYU18} or data selection \citep{DBLP:journals/corr/SchulmanMLJA15,DBLP:conf/icml/WangPMACN20}.

A related application in natural language processing is to train multilingual models \citep{DBLP:journals/corr/abs-1907-05019,DBLP:conf/acl/WangTN20,DBLP:conf/emnlp/WangLT20,DBLP:conf/emnlp/ZhouLLGN21,DBLP:conf/iclr/WangTF021}. For instance, MultiDDS \citep{DBLP:conf/acl/WangTN20} learns a data scorer to balance the data usage of languages. However, designing pre-training is more challenging for lack of prior knowledge, e.g. data size \citep{DBLP:journals/tacl/JohnsonSLKWCTVW17}, data resource \citep{DBLP:conf/emnlp/NeubigH18}. Besides, one can not access to real downstream tasks. All these can lead to so different optimization designs.

\subsection{Meta Learning}

Meta-Learning (Learning to Learn) \citep{DBLP:books/sp/98/ThrunP98} has a long history with vast contributing literature, whereas we could only mention several related works here. \citet{DBLP:conf/iclr/RaviL17} designs an LSTM-based meta-learner to learn the update rule for few shot learning. \citet{DBLP:conf/icml/FinnAL17} proposes MAML to learn an optimized initialization ready for fast adaption to new tasks. The idea also emerges in recent natural language processing, e.g. generating the text mask for MLM \citep{DBLP:conf/emnlp/KangHH20}, optimizing the first-order approximation of dropout to learn dynamic attention pattern \citep{DBLP:journals/corr/abs-2104-04692}, leveraging MAML-inspired pre-training to find a global representation of downstream tasks \citep{DBLP:journals/corr/abs-2004-05568,DBLP:conf/naacl/KeSSMWQ21}.

\section{Multi-Objective Meta-Sampler}

In this section, we first take an overview of our meta-learning framework. What follows is the preliminaries of the pre-training setting as well as a number of ruled-based samplers. Then we discuss the details of our meta-sampler.

\begin{figure*}
\centering
\includegraphics[width=0.99\textwidth]{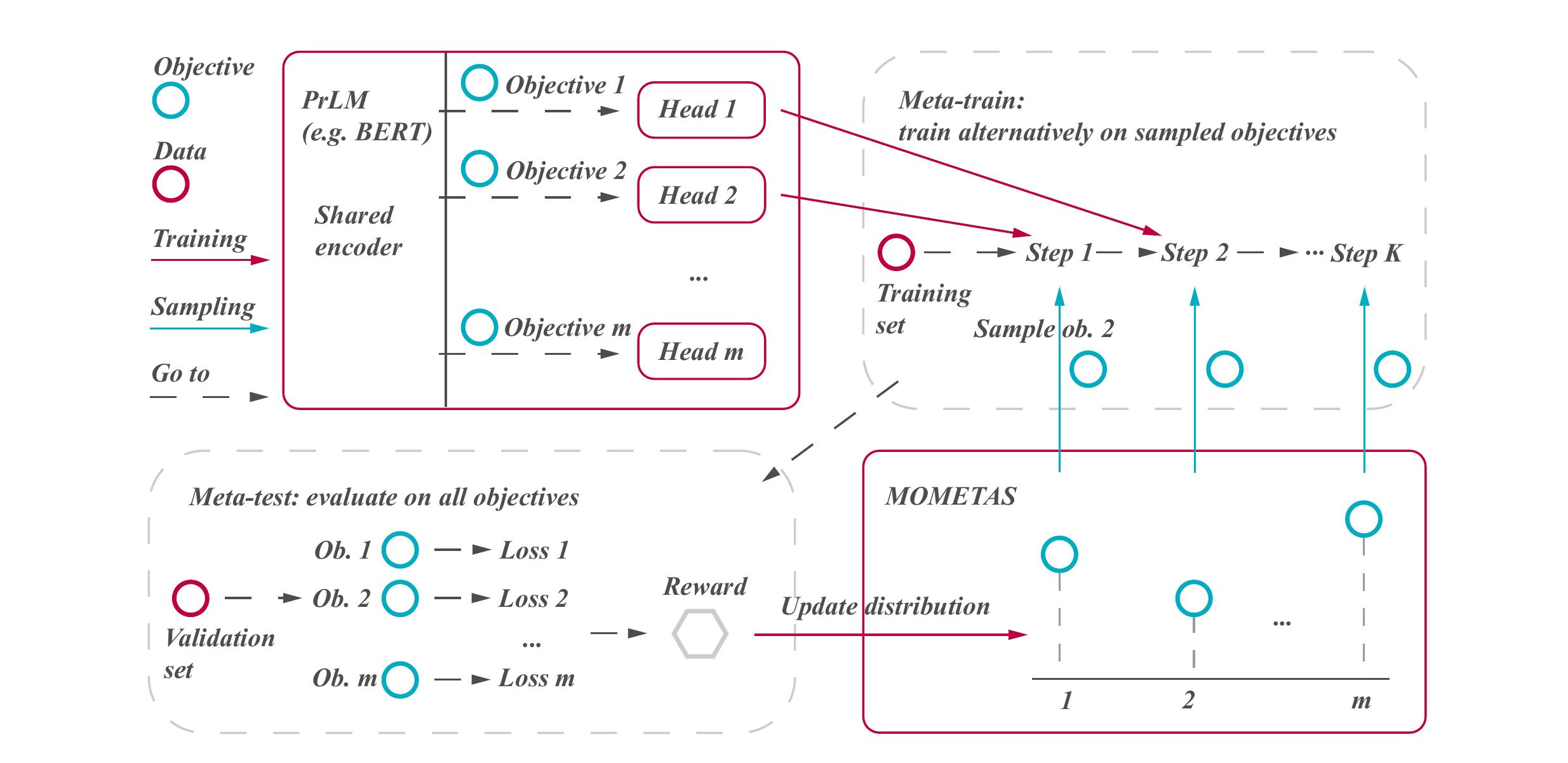}
\caption{An overview of the meta-learning framework of training PrLMs with MOMETAS, where "ob." serves the short for "objective". We only show the first two and the last samplings for simplicity.}
\label{f1}
\end{figure*}

\subsection{Overview}

As depicted in Figure \ref{f1}, we learn the problem in two phases, meta-train and meta-test. In meta-train, the model is trained and updated on a series of pre-training objectives sampled through MOMETAS one by one. After a number of steps, it goes through meta-test, where we evaluate the model over all objectives in one shot. The evaluation is done on a clean validation set in addition to the training one. Based on the evaluation feedback, MOMETAS is then updated. We repeat such train-test cycles until the end of pre-training.

\subsection{Multi-Objective Pre-training}

In our multi-objective pre-training, the model is trained on $ m $ different objectives. The input text of each objectives passes a common encoder to obtain the shared language representation and then output through a specific layer (or head). We denote all objectives as $ \{\mathcal T^1, \mathcal T^2, \cdots, \mathcal T^m\} $, the sampling of which is subject to the latent distribution $ P_{\mathcal D} $. At each training step $ t $, a single objective $ \mathcal T_t \in \{\mathcal T^1, \mathcal T^2, \cdots, \mathcal T^m\} $ is sampled from $ P_{\mathcal D} $.

\subsection{Rule-based Samplers}

We first consider several rule-based samplers:

\noindent $\bullet$ \textit{Uniform-based}: The most straightforward and simplest approach is to make uniform sampling over all objectives. It equals conventional multi-objective training and multi-task learning. However, when the number of objectives is up, it is hard to guarantee the training efficiency, since some simpler objectives come close to convergence early, while some more difficult ones still require a large number of steps to learn well.

\noindent $\bullet$ \textit{Gradient-based}: Gradient acts as a contributing signal of the training state of a network when making gradient descent \citep{DBLP:conf/iclr/RaviL17,DBLP:conf/acl/WangTN20,DBLP:conf/nips/YuK0LHF20}. Larger gradient may have a greater impact on updating its parameters. An intuitive idea is to sample more on those objectives with large gradients, while less on those with small gradients which tend to take minimal impacts on the network. Computationally, we may take the norm of gradients over all encoder parameters \citep{DBLP:conf/iclr/RaviL17}.

\noindent $\bullet$ \textit{Loss-based}: Similar as above, loss acts as another contributing signal of how well a certain objective is learned \citep{DBLP:conf/cvpr/KendallGC18}. More specifically, we may compute the inverse training rate (IR) by dividing the current loss by its initial value, so that lower IR corresponds to a faster training rate for the objective. Thus, the idea is to sample more on those objectives with higher inverse training rates.

\subsection{Meta-Sampler}

Both gradient-based and loss-based approaches merely focus on the state of a single objective in an ad-hoc manner but do not take into account the coupling between them, which makes it hard to achieve the optimal point across all objectives.

Thus, we propose to learn a meta-sampler MOMETAS parametrized as $ \psi=P_{\mathcal D} $, based on meta-learning. Suppose that we sample a single objective at each step $ t $ from $ P_{\mathcal D} $ during meta-train and obtain a sequence of objectives:
$$
\tau = \{\mathcal T_1, \mathcal T_2, \cdots, \mathcal T_K\},\tau \sim P_{\mathcal D}
$$
where $ K $ refers to the number of steps of meta-train (we call it meta length in the paper). In the following meta-test, we evaluate the model over all objectives $ \mathcal T_{1:K} $ on an additional validation set $ \mathcal V $. The goal of MOMETAS is to learn well or earn more gain on all objectives, that is to maximize:
\begin{equation}
J(\psi) = E_{\tau \sim P_{\mathcal D}}[R(\tau)]
\label{e1}
\end{equation}
where $ R(\tau) $ refers to the overall gain given $ \tau $.

Since $ J(\psi) $ is non-differentiable, it is impossible to apply normal gradient-based methods to update MOMETAS which makes sampling from different objectives. Following REINFORCE \citep{DBLP:conf/nips/SuttonMSM99},  we take a number of policy gradient steps to accommodate the non-differentiable operations of sampling, that is:
\begin{equation}
\psi \leftarrow \psi + \beta \sum_{t=1}^{K}\nabla_{\psi}\log{P(\mathcal T_t;\psi)}R(\tau)
\label{e2}
\end{equation}
where $ \beta $ refers to the meta step size. From this perspective, $ R(\tau) $ can be viewed as a rewarding function of training gain. Note that $ R(\tau) $ is only obtained at the end of meta-train ($ t=K $).

Meta length $ K $ indicates the accumulation of meta knowledge. Intuitively, larger $ K $ comes to more training samples until each meta update step, which stabilizes the training process but lowers down the sensitivity of MOMETAS.

\subsubsection{Individual Rewarding}

We further explore the details of the rewarding function $ R(\tau) $. We first let $ r^i $ be the individual gain on each objective ($ i=1 \sim m $) so that $ R(\tau)=\sum_{i=1}^{m}{r^i} $. However, our empirical results show that simply letting $ r^i $ be the opposite of each evaluation loss merely leads to limited performance. This is caused by the problem that it cannot address the issue of negative transfer. Suppose that there is a dominant objective, trained well so that the loss of it is continually down. The real situation can be that the overall loss is declining, while the individual losses of certain objectives are still rising, even though MOMETAS is positively rewarded.

To destroy such confusion, we let $ r^i $ be the \textbf{loss drop} of each objective. Specifically, to compute each loss drop, we always maintain the last loss value as the baseline $ b^i $ (the evaluation loss from last meta-test). Then we compare the current loss value $ a^i $ (from current meta-test) with it. Because the magnitude of loss differs from objectives, we further compute the relative loss drop by dividing it by the baseline $ b^i $. Hence, the final rewarding function can be formulated as:
\begin{equation}
R(\tau) = \sum_{i=1}^{m}{\frac{b^i-a^i}{b^i}}
\label{e3}
\end{equation}
where $ b^i $ and $ a^i $ refer to the loss values of the last meta-test and current meta-test respectively. Such rewarding function forces MOMETAS to explore the optimal sampling pattern which is useful across all pre-training objectives.

\subsubsection{Entropy Regularization}

To further escape from the local optimum, we impose maximum entropy regularization as an additional constraint \citep{DBLP:conf/icml/HaarnojaZAL18}, which is widely used in stochastic reinforcement learning. The idea behind this is that smaller entropy means more deterministic sampling from the distribution and MOMETAS will be punished in this situation, which encourages MOMETAS to explore and allows it to step out of the local optimal point. Hence, the training objective of MOMETAS comes to:
\begin{equation}
J(\psi) = E_{\tau \sim P_{\mathcal D}}[R(\tau) + \lambda H(\psi)]
\label{e4}
\end{equation}
where $ H(\psi) $ refers to the entropy regularization term. We find good performances when the temperature parameter $ \lambda $ is set to $ 1 \sim 3 $.

\begin{algorithm}[tb]
\caption{Pre-train with MOMETAS}
\textbf{Input:} Model $ \theta $, $ m $ pre-training objectives $ \{\mathcal T^1, \mathcal T^2, \cdots, \mathcal T^m\} $, meta length $ K $, MOMETAS distribution $ P_{\mathcal D} $, validation set $ \mathcal V $
\begin{algorithmic}[1]
\State{Initialize $ \mathcal D $ with uniform distribution}
\While {not converged}
    \State{Empty $ \tau $}
    \For {$ t=1 $ to $ K $}
      \State {Sample one objective $ \mathcal T_t \sim P_{\mathcal D} $}
      \State {Update model parameters $ \theta_t $}
      \State {Append $ \mathcal T_t $ into $ \tau $}
    \EndFor
    \State{Fetch data for each objective from $ \mathcal V $}
    \State{Evaluate with model parameters $ \theta_K $}
    \State{Compute reward via Eq. \ref{e3}}
    \State{Update $ P_{\mathcal D} $ via Eq. \ref{e2}}
\EndWhile
\end{algorithmic}
\label{a1}
\end{algorithm}

\begin{table*}[t]
\centering
\begin{tabular}{@{}lccccccccc@{}}
\toprule
         & \begin{tabular}[c]{@{}c@{}}CoLA\\ (Mcc)\end{tabular} & \begin{tabular}[c]{@{}c@{}}SST-2\\ (Acc)\end{tabular} & \begin{tabular}[c]{@{}c@{}}MRPC\\ (Acc)\end{tabular} & \begin{tabular}[c]{@{}c@{}}QNLI\\ (Acc)\end{tabular} & \begin{tabular}[c]{@{}c@{}}MNLI-m/mm\\ (Acc)\end{tabular} & \begin{tabular}[c]{@{}c@{}}QQP\\ (F1)\end{tabular} & \begin{tabular}[c]{@{}c@{}}RTE\\ (Acc)\end{tabular} & \begin{tabular}[c]{@{}c@{}}STS-B\\ (Spc)\end{tabular} & Avg \\ \midrule
BERT$_{base}$                     & 51.9               & 93.5               & 88.9               & 90.5               & 84.6/83.4                & 71.2                & 66.4                & 85.8                & 79.6 \\ \hdashline
BERT$_{base}$ (Ours)              & 52.1               & 92.9               & 88.7               & 90.2               & 84.6/83.4                & 71.3                & 67.4                & 84.6                & 79.5 \\ \hdashline
\quad + \textit{Ub}      & 52.0               & 93.0               & 89.1               & 90.6               & 84.7/83.7                & 71.5                & 66.7                & 85.0                & 79.7 \\
\quad + \textit{Gb}      & 52.0               & 93.6               & 89.2               & \textbf{90.7}      & 84.5/84.0                & 71.8                & 66.9                & 85.9                & 79.8 \\
\quad + \textit{Lb}      & 53.1               & 93.3               & 89.7               & 90.5               & 84.8/\textbf{84.4}       & 71.8                & 67.3                & 86.0                & 80.1 \\
\quad + \textit{MOMETAS} & \textbf{55.9}      & \textbf{93.7}      & \textbf{90.0}      & \textbf{90.7}      & \textbf{85.2}/84.3       & \textbf{72.1}       & \textbf{68.4}       & \textbf{86.9}       & \textbf{80.8} \\ \bottomrule
\end{tabular}
\caption{GLUE test results under different sampling strategies. BERT$_{base}$ refers to the reported results in \citet{DBLP:conf/naacl/DevlinCLT19} while BERT$_{base}$ (Ours) refers to our rerun results. Due to limited number of submissions per day, we do not report the results over multiple runs in Table \ref{t1} (for multiple runs, please refer to Table \ref{t2}).}
\label{t1}
\end{table*}

\subsubsection{Algorithm}

Then we present our meta-learning algorithm, which is summarized in Algorithm \ref{a1}. Specifically, we first initialize MOMETAS distribution $ P_{\mathcal D} $ with uniform distribution. In meta-train, the model is fed with $ K $ sampled pre-training objectives one by one. At each step $ t $, we need to record every single sampling $ \mathcal T_t $ in order to update MOMETAS later. What follows is meta-test, where the model is evaluated on the validation set $ \mathcal V $. MOMETAS will be rewarded based on the evaluation feedback and then updated so as to be ready for the next meta-train. We repeat such a train-test cycle for times until model convergence. Note that we fetch the validation samples from $ \mathcal V $ through random sampling to guarantee the training efficiency.

When pre-training with MOMETAS, the additional time consumption mainly comes from doing evaluation in meta-test. Though it will rise as the number of objectives increases, the evaluation is done only once every $ K $ steps (e.g. 100) and is inherently fast with no backward passes. Thus, the overhead brought by MOMETAS is minimal.

\section{Experimental Setup}

In this section, we present our experimental setup. Our implementations are based on PyTorch using \emph{transformers} \citep{DBLP:conf/emnlp/WolfDSCDMCRLFDS20}.

\subsection{Pre-training Objectives}

We adopt five pre-training objectives in our experiments. The details of them are listed below.

\noindent $\bullet$ \textit{General Language Representation} - \textbf{Masked Language Modeling (MLM)}: Following BERT \citep{DBLP:conf/naacl/DevlinCLT19}, we randomly sample 15\% of the tokens in each input sequence and replace them with special \texttt{[MASK]} elements.

\noindent $\bullet$ \textit{Semantic} - \textbf{Contrastive Learning of Sentence Embeddings (CSE)}: Following SimCSE \citep{DBLP:conf/emnlp/GaoYC21}, we feed the same sequence twice by applying different dropout masks and extract the \texttt{[CLS]} elements as their sentence representations. The model is required to predict the input sentence itself from in-batch negatives.

\noindent $\bullet$ \textit{Coherence} - \textbf{Added Token Detection (ATD)}: We randomly sample 15\% of the positions in each sequence and insert random tokens in them. The model is required to decide which positions are superfluous. Different from MLM, ATD expands the context of text. Different from MLM, ATD expands the context of text. \textbf{Deleted Token Detection (DTD)}: Similar as ATD, we randomly remove those tokens and the model is required to decide which positions are missed.

\noindent $\bullet$ \textit{Entity \& Knowledge} - \textbf{Entity-guided Masked Language Modeling (EMLM)}: We leverage the prior knowledge to further strengthen the model. We first pick out the entities in each sequence\footnote{\url{https://github.com/stanfordnlp/stanza}} and then randomly replace a half of them with \texttt{[MASK]} \citep{DBLP:conf/iclr/XiongDWS20}.

Though we are unable to cover all alternatives in this paper, the experiments are of great potential to be extended to other pre-training setups.

\begin{table*}[t]
\centering
\resizebox{0.99\textwidth}{!}{\begin{tabular}{@{}lccccccccccccc@{}}
\multicolumn{12}{c}{} \\ \toprule
\multirow{3}{*}{\textbf{Model}}
&  \multicolumn{2}{c}{\small \textit{\textbf{Language Inference}}}
&& \multicolumn{2}{c}{\small \textit{\textbf{Semantic Similarity}}}
&& \multicolumn{2}{c}{\small \textit{\textbf{NER}}}
&& \multicolumn{2}{c}{\small \textit{\textbf{Multi-Choice MRC}}} \\
\cline{2-3} \cline{5-6} \cline{8-9} \cline{11-12} \rule{0pt}{12pt}
                 & MNLI                   & SICK                   && P-QQP                   & STS-B                  && CoNLL                  & WNUT                    && DREAM                   & aNLI                    \\
                 & (Acc)                  & (Acc)                  && (Acc)                   & (Spc)                  && (F1)                   & (F1)                    && (Acc)                   & (Acc)                   \\ \midrule
\specialrule{0em}{.0pt}{.0pt}
BERT$_{base}$    & 83.9                   & 87.0                   && 33.4$_{(0.6)}$          & 84.8                   && 91.2                   & 48.8$_{(1.0)}$          && 62.5$_{(0.6)}$          & 63.8$_{(0.5)}$          \\ \hdashline
\multicolumn{12}{c}{\small \textit{Single-objective pre-training}} \\ \hdashline
\textit{MLM}     & \textbf{84.6} 	      & \textbf{87.4} 	       && 34.1$_{(0.6)}$ 	      & 84.5 	               && 91.4 	                 & 51.8$_{(0.4)}$ 	       && 62.5$_{(0.3)}$ 	      & 64.0$_{(0.2)}$          \\
\textit{CSE}     & 83.6                   & 85.8                   && 32.1$_{(0.9)}$          & \textbf{86.1}          && 91.0                   & 46.3$_{(0.8)}$          && 53.2$_{(1.8)}$          & 63.6$_{(0.6)}$          \\
\textit{ATD}     & 84.3                   & \textbf{87.4}          && 34.3$_{(0.5)}$          & 84.6                   && 91.4                   & 48.8$_{(0.9)}$          && 57.3$_{(1.1)}$          & \textbf{64.5}$_{(0.9)}$ \\
\textit{DTD}     & 83.9                   & 86.3                   && \textbf{36.2}$_{(1.2)}$ & 85.0                   && 91.6                   & 50.5$_{(1.5)}$          && 58.7$_{(0.8)}$          & 62.9$_{(0.3)}$          \\
\textit{EMLM}    & 84.3                   & 86.6                   && 34.4$_{(1.0)}$          & 85.8                   && \textbf{92.1}          & \textbf{53.1}$_{(0.6)}$ && 60.2$_{(0.8)}$          & 63.1$_{(0.3)}$          \\ \hdashline
\multicolumn{12}{c}{\small \textit{Multi-objective pre-training}} \\ \hdashline
\textit{Ub}      & 84.2                   & 87.5                   && 35.6$_{(0.8)}$          & 85.2                   && 91.6                   & 50.8$_{(0.7)}$          && 63.2$_{(0.5)}$          & 64.6$_{(0.8)}$          \\
\textit{Lb}      & 84.6                   & 87.3                   && 33.1$_{(0.9)}$          & 86.0                   && 91.7                   & 50.7$_{(0.9)}$          && \textbf{64.6}$_{(0.3)}$ & 65.2$_{(0.2)}$          \\
\textit{MOMETAS} & \textbf{84.8}          & \textbf{87.9}          && \textbf{36.5}$_{(0.4)}$ & \textbf{86.5}          && \textbf{92.0}          & \textbf{52.1}$_{(1.0)}$ && 64.5$_{(0.3)}$          & \textbf{65.8}$_{(0.3)}$ \\ \midrule
\specialrule{0em}{.0pt}{.0pt}
BERT$_{large}$   & 86.1                   & 87.6                   && 36.2$_{(0.9)}$          & 86.4                   && 91.9                   & 50.2$_{(1.5)}$          && 66.3$_{(1.3)}$          & 66.9$_{(0.8)}$          \\ \hdashline
\multicolumn{12}{c}{\small \textit{Single-objective pre-training}} \\ \hdashline
\textit{Ub}      & 86.1                   & 88.2                   && 40.6$_{(0.5)}$          & 87.5                   && 92.3                   & 50.9$_{(1.8)}$          && 65.8$_{(0.8)}$          & 67.7$_{(0.7)}$          \\
\textit{MOMETAS} & \textbf{86.5}          & \textbf{88.6}          && \textbf{41.8}$_{(0.5)}$ & \textbf{88.5}          && \textbf{92.4}          & \textbf{52.9}$_{(1.2)}$ && \textbf{68.5}$_{(0.7)}$ & \textbf{69.1}$_{(0.5)}$ \\ \bottomrule
\end{tabular}}
\caption{Results on more tasks over five runs. We report the mean and standard deviation. Respectively, \textit{Base} refers to the rerun original BERT model, and \textit{Ub}, \textit{Lb}, and \textit{MOMETAS} refer to the multi-objective trained models based on corresponding sampling strategies. For MNLI, we average the two scores of \textit{m} and \textit{mm} divisions.}
\label{t2}
\end{table*}

\subsection{Dataset}

Based on our pre-training setup, we validate our approach on a wide range of downstream benchmarks (14 tasks in total). In what follows, we summarize them as well as describe how the chosen ones relate to our pre-training objectives.

\paragraph{General Natural Language Understanding} We adopt \textbf{GLUE} benchmark \citep{DBLP:conf/iclr/WangSMHLB19}, a collection of eight natural language understanding tasks, including natural language inference, sentiment analysis and semantic similarity. We exclude problematic WNLI as in \citet{DBLP:conf/naacl/DevlinCLT19}). In addition, we adopt \textbf{SICK} \citep{DBLP:conf/lrec/MarelliMBBBZ14}, another natural language inference benchmark as a complement.

\paragraph{Semantic Similarity} We further adopt \textbf{PAWS-QQP} \citep{DBLP:conf/naacl/ZhangBH19}, which adds adversarial examples to QQP for evaluating model robustness. Following the zero-shot setting in \citet{DBLP:conf/naacl/ZhangBH19}, we train the model on QQP and directly evaluate it on PAWS-QQP.

\paragraph{Named Entity Recognition (NER)} We adopt two benchmarks, \textbf{CoNLL-2003} \citep{DBLP:conf/conll/SangM03} and \textbf{WNUT-2017} \citep{DBLP:conf/aclnut/DerczynskiNEL17}. Of these, WNUT-2017 contains a large number of rare entities, which therefore requires the model with stronger generalization.

\paragraph{Multi-choice Machine Reading Comprehension (MRC)} Two challenging benchmarks are adopted, \textbf{DREAM} \citep{DBLP:journals/tacl/SunYCYCC19} for multi-turn dialogue understanding, and \textbf{aNLI} \citep{DBLP:conf/iclr/BhagavatulaBMSH20} for commonsense reasoning, both of which are in format of multi-choice MRC.

Notably for DREAM and aNLI, there are no straightforward objectives adopted. However, it is desirable that the model is able to learn the interdisciplinary knowledge and generalize better on tasks not seen during pre-training through jointly learning multiple objectives.

\subsection{Baseline Strategies}

We compare MOMETAS with several earlier discussed sampling strategies, including \textit{Uniform-based} (Ub), \textit{Gradient-based} (Gb), and \textit{Loss-based} (Lb). Experiments are made on BERT$_{base}$ models.

Except for Ub, the rest two are based on proportion, that is we sample the objectives as proportional to the magnitudes of concerned values. To implement, we compute the average gradient (L2 norm of gradients over encoder parameters) or loss of each objective for every certain number of training steps (to keep in pace with MOMETAS, also $ K $ steps). At the same point as meta-test, we update the distribution. However, we find some large values (e.g. big gradient at the start of training) will make the probabilities of other objectives close to zero. Following \citet{DBLP:conf/nips/AndrychowiczDCH16}, we use Sigmoid function to scale them properly.

\subsection{Training Details}

\paragraph{Pre-training} Inherited from the released checkpoints, \texttt{bert-base-uncased} and \texttt{bert-large-uncased}\footnote{\url{https://github.com/huggingface/transformers/}}, we continue to pre-train our models following multi-objective setting. For training corpus, we use a subset of Colossal Clean Crawled Corpus \citep{DBLP:journals/jmlr/RaffelSRLNMZLL20} (we use nearly 100GB of it and randomly sample 1GB for validation). Each single model is trained with 512 batch size and for 50K steps (nearly one epoch). Unless otherwise specified, we fix meta length $ K $ to 100 and meta step size to 1e-1. Training a base/large-size model takes about 12/36 hours on 8 V100 GPUs with FP16 for both uniform-based sampling and MOMETAS.

\paragraph{Fine-tuning} For all GLUE sub-tasks, we follow the hyperparameters shared in \citet{DBLP:conf/iclr/LanCGGSS20} and fine-tune for 3 epochs, except 10 epochs for RTE and STS-B. For other tasks, we merely sweep through learning rates and batch sizes for efficiency, excluding dropout probabilities or weight decay rates. Readers can refer to Appendix \ref{ap1} for details.

\section{Empirical Results}

\paragraph{GLUE} Table \ref{t1} reports the test results on GLUE benchmark under different sampling strategies, all of which are based BERT$_{base}$. Intuitively, simple uniform multi-objective pre-training (Ub) merely leads to limited performance gain (79.5 $\rightarrow$ 79.7). Besides, we find that Gb is also not effective, while Lb brings nice gain (\textbf{79.5} $\rightarrow$ \textbf{80.1}). However, more powerful performance gain can be seen on MOMETAS-empowered one (\textbf{79.5} $\rightarrow$ \textbf{80.8}). Compared to Ub, MOMETAS outperforms it on all eight sub-tasks (\textbf{3.9} points absolute gain on CoLA, \textbf{0.7} on SST-2, \textbf{0.9} on MRPC, \textbf{1.7} on RTE, \textbf{2.3} on STS-B), which indicates the strength of our meta-learning-based sampling.

\paragraph{More tasks} We make further experiments on more different tasks as in Table \ref{t2}. Generally, MOMETAS better facilitates multi-objective pre-training compared to Ub and Lb. We first focus on two semantic similarity tasks (STS-B and PAWS-QQP), for which we adopt CSE to improve the performance. According to \citet{DBLP:conf/emnlp/GaoYC21}, single CSE-trained BERT can achieve significant improvement. When the number of objectives increases, however, the situation can be difficult. It does not work well with Ub (84.8 $\rightarrow$ 85.2 on STS-B). Contrarily, MOMETAS brings a huge performance boost on BERT$_{base}$ (\textbf{84.8} $\rightarrow$ \textbf{86.5} on STS-B, \textbf{33.4} $\rightarrow$ \textbf{36.5} on P-QQP), even surpasses BERT$_{large}$. Similar situation can be found on NER comparing Ub with MOMETAS (\textbf{50.8} $\rightarrow$ \textbf{52.1} on WNUT). It demonstrates that \textbf{MOMETAS helps maintain the benefit of a single objective in the multi-objective scenario}. Additionally, MOMETAS-empowered BERT$_{base}$ is able to outperform BERT$_{large}$ on some of the tasks (SICK, P-QQP, STS-B, CoNLL and WNUT), suggesting the great potential of multi-objective pre-training. On the other hand, because of the attempt to learning cross knowledge from other objectives, MOMETAS also enables the model to learn well on MRC tasks, even though there are no related objectives adopted.

\paragraph{Single-objective} Table \ref{t2} also demonstrates the superiority of multi-objective pre-training over single-objective. We find that the overall performance gain brought by each single objective is limited. Though it may lead to notable improvement on certain tasks (e.g. EMLM on NER), it may also cause the model to perform particularly badly on others (CSE on NER). However, MOMETAS is designed to find the all-round direction where the model is able to perform well on all objectives.

\section{Visualization}

\begin{figure}
\centering
\includegraphics[width=0.43\textwidth]{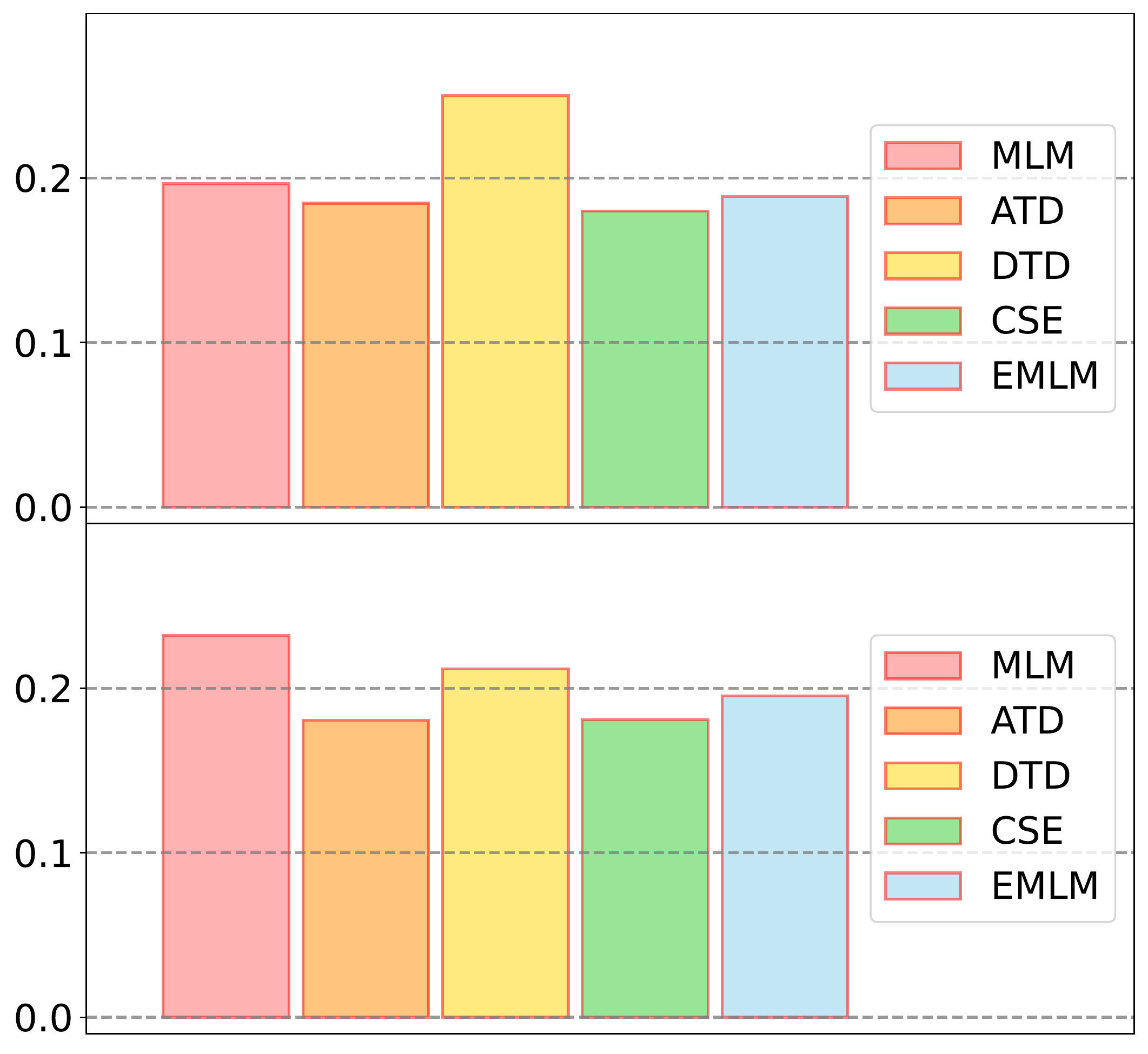}
\caption{Averaged sampling weights learned by MOMETAS, upper for $ \lambda=2 $, bottom for $ \lambda =3 $.}
\label{f2}
\end{figure}

\paragraph{Probability distribution} Figure \ref{f2} depicts the sampling weighs averaged throughout the training process of all pre-training objectives learned by MOMETAS. Intuitively, the distribution looks more volatile when $ \lambda=2 $ (upper), while more clustered when $ \lambda=3 $ (bottom), which indicates the role of entropy regularization. From both cases, we may find some common clues. DTD always stands a high picking weight, which uncovers the potential of deleting corruption when learning a denoising encoder. In addition, both MLM and EMLM are never underweight, and a general masking strategy outweighs a specific one. Then we look at CSE, a sentence-level objective, which is much easier than the other token-level ones. We find its picking weight is very high in the early period of training but drops quickly in the later period.

\begin{figure}
\centering
\includegraphics[width=0.45\textwidth]{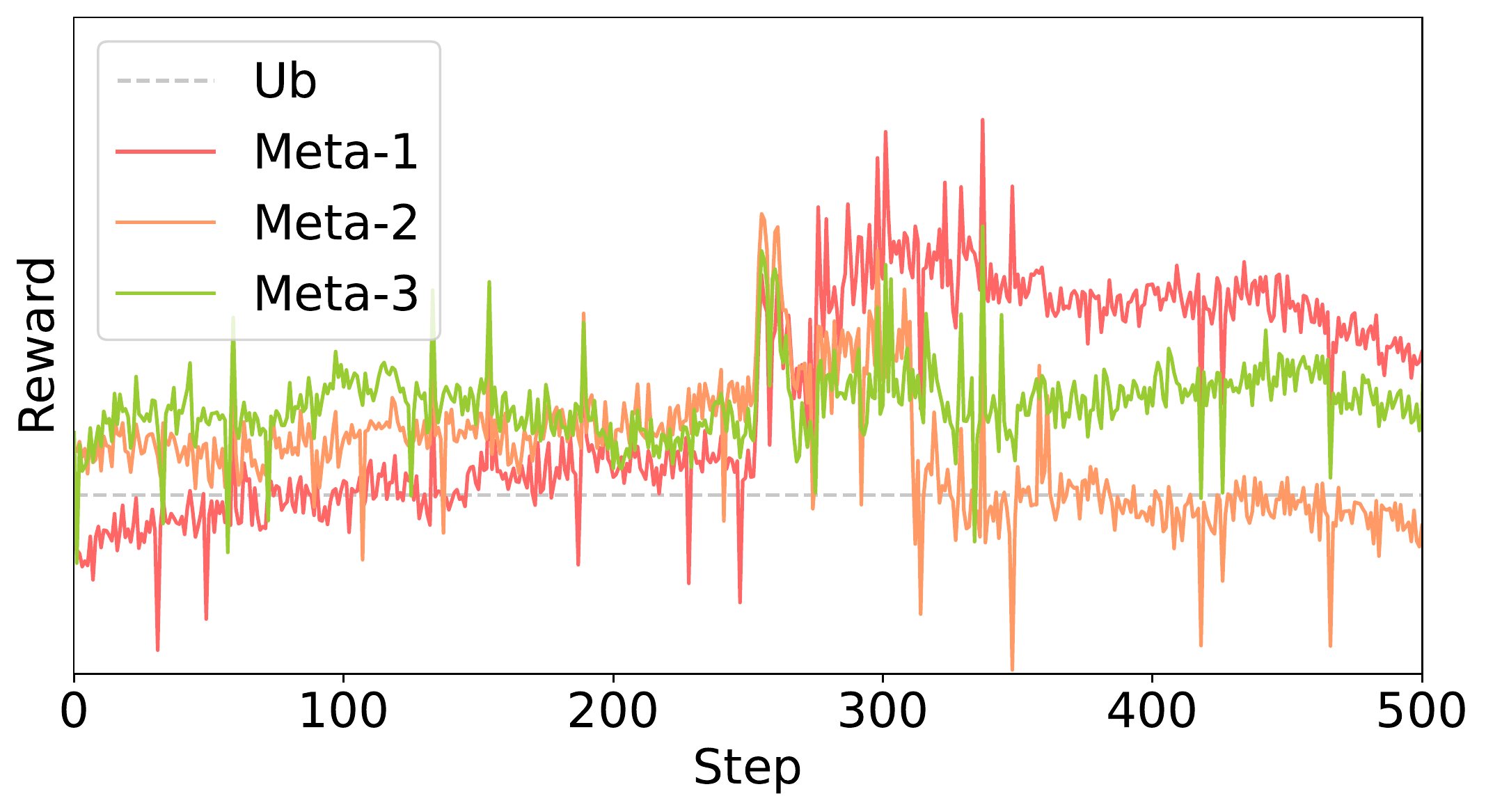}
\caption{Difference of the total reward, where Ub (a horizontal line of 0) and Meta-$ x $ refer to the uniform-based sampling and MOMETAS with entropy regularization $ \lambda=x $. To make more intuitive, we smooth the curves by convolution.}
\label{f3}
\end{figure}

\paragraph{Reward} We observe the respective reward curves of MOMETAS and Ub to access to their training gain for multi-objective pre-training. To make intuitive, we depict the difference of them (the former minus the latter) as in Figure \ref{f3}. Intuitively, we see slight differences at the beginning of training since MOMETAS is initialized with uniform distribution. However, all three curves are positive for majority of the time.  When $ \lambda=1 $ for instance, we see a rising trend of the curve, from negative to positive, while when $ \lambda=3 $, the curve is always above zero, which implies that MOMETAS learns to achieve more evaluation scores than Ub in meta-test.

\section{Ablation Studies}

This section reports our ablation studies over a number of factors of MOMETAS in order to better understand their roles. For all experiments, we report the results over five runs.

\begin{table}[]
\centering
\begin{tabular}{@{}lccc@{}}
\toprule
                         & SICK                   & STS-B                  & WNUT                    \\ \midrule
\textit{Overall}         & 87.2$_{(.2)}$          & 85.2$_{(.5)}$          & 50.0$_{(.8)}$           \\
\textit{Hard indiv.}     & 87.6$_{(.0)}$          & 86.2$_{(.2)}$          & 51.0$_{(1.1)}$          \\
\textit{Relative indiv.} & \textbf{87.9}$_{(.3)}$ & \textbf{86.5}$_{(.2)}$ & \textbf{52.1}$_{(.6)}$ \\ \bottomrule
\end{tabular}
\caption{Comparison between rewarding functions of MOMETAS on BERT$_{base}$. We keep $ K $ and $ \lambda $ the same.}
\label{t3}
\end{table}

\subsection{Comparison between Rewarding Functions}

We compare different rewarding functions $ R(\tau) $ on three GLUE sub-tasks, SST-2, QNLI and STS-B: (1) \textbf{overall loss rewarding}: we optimize the summation of all losses; (2) \textbf{relative individual rewarding}: exactly what we use in MOMETAS, we optimize the summation of all relative loss drops as Eq. \ref{e3}; (3) \textbf{hard individual rewarding}: similar as the relative one, we replace the individual loss drop with $ \pm 1 $ when it is down or up respectively and optimize the summation of them.

As shown in Table \ref{t3}, slight improvement can be seen when simply rewarding MOMETAS with overall loss compared to uniform-based sampling in Table \ref{t1}. In this situation, it is hard to learn the balance between all objectives. However, individual rewarding can achieve stronger performances in both hard and relative cases.

\subsection{Effect of Entropy Regularization}

When optimizing MOMETAS, we apply maximum entropy regularization to encourage exploration in the hope of seeking out the global optima. Table \ref{t4} demonstrates the effect of different degrees of entropy regularization on pre-training performances. We can see general gain compared to original BERT in Table \ref{t1} even if there is no regularization applied. However, regularization further boosts the performances. The best case occurs when $ \lambda=3 $, which the model outperforms the base one by 0.5, 0.7 and 1.1 points on all three tasks, respectively.

\begin{table}[]
\centering
\begin{tabular}{@{}lccc@{}}
\toprule
                                & MNLI-m                 & STS-B                  & WNUT                   \\ \midrule
\textit{Base} ($ \lambda = 0 $) & 84.7$_{(.0)}$          & 85.8$_{(.3)}$          & 51.0$_{(.6)}$          \\
$ \lambda = 1 $                 & 85.1$_{(.1)}$          & 86.2$_{(.2)}$          & 51.7$_{(.6)}$          \\
$ \lambda = 2 $                 & \textbf{85.3}$_{(.2)}$ & 86.2$_{(.5)}$          & 50.8$_{(.2)}$          \\
$ \lambda = 3 $                 & 85.2$_{(.2)}$          & \textbf{86.5}$_{(.2)}$ & \textbf{52.1}$_{(.7)}$ \\ \bottomrule
\end{tabular}
\caption{Effect of entropy regularization on BERT$_{base}$. The base model is trained with no regularization.}
\label{t4}
\end{table}

\subsection{Effect of Meta Length}

In our pre-training framework, MOMETAS is designed to be updated every $ K $ steps. $ K $ refers to the number of steps of meta-train and meanwhile reflects the knowledge accumulation before meta-test. Generally, when $ K $ becomes larger, MOMETAS tends to be less sensitive and pay more attention to long-term benefits. Contrarily, when $ K $ is close to 1, it is greedy and only cares about the current moment. In practical, it cannot be smaller than the number of objectives.

Table \ref{t5} shows the pre-training performances under a number of values of $ K $. We can see that a too small $ K $ may lead to worse results (e.g. $ K=25 $). It can be presumed that \textbf{long-sight helps to find the global optimum}. For example, we cannot acquire sufficient meta knowledge to justify all objectives when $ K $ is too small. This can be supported by another fact that \textbf{MOMETAS is found more uniform-distributed when $ K $ becomes smaller} under the same degree of entropy regularization. On the other hand, we can see nice results when $ K $ is larger (e.g. $ K=100,200 $). It hints that we can choose a properly larger $ K $ to speedup the pre-training since there will be less meta-test steps.

\begin{table}[]
\centering
\begin{tabular}{@{}lcccc@{}}
\toprule
        & MNLI-m        & SICK          & WNUT            & Avg           \\ \midrule
$K=25$  & 84.6          & 87.5          & 51.3            & 74.5          \\
$K=50$  & 85.1          & 87.5          & 51.7            & 74.8          \\
$K=100$ & \textbf{85.2} & \textbf{87.9} & 52.1            & \textbf{75.1} \\
$K=200$ & 85.0          & 87.7          & \textbf{52.4}   & 75.0          \\ \bottomrule
\end{tabular}
\caption{Effect of meta length on BERT$_{base}$. Note that the results are based on five runs but we do not list the variances for space limitation.}
\label{t5}
\end{table}

\section{Conclusion}

This paper concentrates on multi-objective pre-training of PrLMs and presents Multi-Objective Meta-Sampler (MOMETAS) in the hope of combining arbitrary pre-training objectives organically. We adopt five pre-training objectives and conduct experiments on the base-size and large-size models. The empirical results on a wide range of NLP tasks demonstrate that MOMETAS largely outperforms other rule-based sampling strategies and unlocks more powerful language models.

\section*{Limitations}

This paper proposes a novel pre-training framework, and therefore requires larger GPU resources. However, we will release our trained checkpoints, pre-training corpus, and code to facilitate further research. Our pre-training experiments are limited in continual pre-training as there are only 8 GPUs available. We therefore expect future researchers to practice and validate our approach when pre-training from scratch, even on stronger model architectures.

This paper discusses fewer on how we choose each single pre-training objective. We do not rule out other potential options that can make the pre-trained model even better. Of course, we will keep following up on this part of the study and train new models. Besides, we do not make the experiments with a larger number of objectives (e.g. 10). It is possible that the optimized value of $ \lambda $ for entropy regularization will be different when the number becomes larger.

Another limitation is that we do not discuss the role of the validation set which is necessary for meta-learning. Intuitively, a carefully-selected validation set may improve the credibility of meta-test. For example, it can be positive to introduce signals that are more related to the downstream tasks. We will leave this part for our future work.


\bibliography{anthology,custom}
\bibliographystyle{acl_natbib}

\appendix

\section{Training Details}
\label{ap1}

\begin{table*}[ht]
\centering
\begin{tabular}{@{}lcc@{}}
\toprule
                          & BERT$_{base}$ & BERT$_{large}$ \\ \midrule
Number of hidden layers   & 12            & 24             \\
Hidden size               & 768           & 1024           \\
Intermediate size         & 3072          & 4096           \\
Number of attention heads & 12            & 16             \\
Dropout                   & 0.1           & 0.1            \\
Batch size                & 512           & 512            \\
Learning rate             & 5e-5          & 5e-5           \\
Weight Decay              & 0.01          & 0.01           \\
Max sequence length       & 256           & 256            \\
Warmup proportion         & 0.06          & 0.06           \\
Max steps                 & 50K           & 50K            \\
Gradient clipping         & 1.0           & 1.0            \\
FP16                      & Yes           & Yes            \\
Number of GPUs            & 8             & 8              \\
Training period           & 12 hours      & 36 hours       \\ \bottomrule
\end{tabular}
\caption{Hyperparameters for pre-training.}
\end{table*}

\begin{table*}[hb]
\centering
\begin{tabular}{@{}lcccccccc@{}}
\toprule
                          & MNLI & SICK & QQP  & STS-B & CoNLL    & WNUT    & DREAM & aNLI \\ \midrule
Dropout                   & 0.1  & 0.1  & 0.1  & 0.1   & 0.1      & 0.1     & 0.1   & 0.1  \\
Batch size                & 128  & 32   & 128  & 16    & 32       & 16      & 16    & 64   \\
Learning rate             & 3e-5 & 5e-5 & 5e-5 & 5e-5  & 5e-5     & 5e-5    & 3e-5  & 5e-5 \\
Weight Decay              & 0.01 & 0.01 & 0.01 & 0.01  & 0.01     & 0.01    & 0.01  & 0.01 \\
Max sequence length       & 128  & 128  & 128  & 128   & 128      & 64      & 128   & 128  \\
Warmup proportion         & 0.06 & 0.06 & 0.06 & 0.06  & 0.1      & 0.1     & 0.06  & 0.06 \\
Max epochs                & 3    & 3    & 3    & 10    & 3        & 5       & 6     & 3    \\
FP16                      & Yes  & Yes  & Yes  & Yes   & Yes      & Yes     & Yes   & Yes  \\ \bottomrule
\end{tabular}
\caption{Hyperparameters for fine-tuning.}
\end{table*}

\end{document}